\pdfoutput=1
\documentclass[11pt]{article}

\usepackage[]{ACL2023}

\usepackage{times}
\usepackage{latexsym}

\usepackage[T1]{fontenc}
\usepackage[utf8]{inputenc}
\usepackage{microtype}
\usepackage{inconsolata}
\usepackage{subcaption}

\usepackage{CJKutf8}

\usepackage{graphicx}
\usepackage{tcolorbox}
\usepackage{paralist}

\usepackage{amsmath,amssymb,amsfonts}

\usepackage{multirow}
\usepackage{booktabs}
\usepackage{arydshln}
\usepackage{bm}

\usepackage{textcomp}

\usepackage{colortbl}

\usepackage{color,xcolor}

\usepackage{tikz}

\definecolor{nmblue}{RGB}{209,242,255}
\definecolor{nmtextblue}{RGB}{116,214,252}
\definecolor{nmpurple}{RGB}{216,216,255}%203,203,245
\definecolor{nmtextpurple}{RGB}{165,124,224}

\definecolor{nmred}{RGB}{255,229,229}
\definecolor{nmdeepred}{RGB}{255,183,183}
\definecolor{nmgreen}{RGB}{196,237,227}%165,232,211
\definecolor{nmdeepgreen}{RGB}{126,227,203}
\definecolor{nmyellow}{RGB}{255,242,216}%250,229,191
\definecolor{nmdeepyellow}{RGB}{250,224,167}

\definecolor{nmgray}{RGB}{229,229,229}

\definecolor{greeni}{RGB}{226,239,217}
\definecolor{yellowi}{RGB}{255,242,204}
\definecolor{purplei}{RGB}{234,193,222}

\definecolor{mygray1}{gray}{0.84}
\definecolor{mygray2}{gray}{0.56}
\definecolor{mygray3}{gray}{0.4}

\definecolor{bluei}{RGB}{192,232,242}
\definecolor{redi}{RGB}{255,184,184}
\definecolor{blueii}{RGB}{207,238,245}
\definecolor{redii}{RGB}{250,241,240}
\definecolor{lightblue}{RGB}{126,126,250}
\definecolor{lightred}{RGB}{255,139,139}
\definecolor{highlightblue}{RGB}{39,39,191}
\definecolor{highlightred}{RGB}{214,13,13}
\definecolor{bgred}{RGB}{234,67,67}
\definecolor{myellowx}{RGB}{255,240,134}
\definecolor{mredx}{RGB}{232,124,108}

\definecolor{highlightgreen1}{RGB}{229,250,245}
\definecolor{highlightgreen2}{RGB}{0,120,87}
\definecolor{purple1}{RGB}{175,20,241}
\definecolor{green1}{RGB}{0,176,80}

\definecolor{nmdeeppurple}{RGB}{170,170,250}
 
\definecolor{nmdeepblue}{RGB}{165,222,237}
\definecolor{nmblue}{RGB}{196,229,237}
 
\definecolor{nmdeeppink}{RGB}{242,178,134}
\definecolor{nmpink}{RGB}{255,214,183}

\tcbuselibrary{most}

\usepackage{adjustbox}

\usepackage{arydshln}

\definecolor{mypurplex}{RGB}{173,173,173}

\usepackage{algorithm}
\usepackage{algorithmic}
\newtcolorbox{mybox}[2][]{
colback = red!75!white, colframe = red!75!black, fonttitle = \bfseries,
colbacktitle = bgred, enhanced,
boxsep=0pt,left=3pt,right=3pt,top=10pt,bottom=3pt,
fontupper=\linespread{0.9}\selectfont,
attach boxed title to top center={yshift=-3mm},
title=#2,#1}

\newenvironment{myquote}%
  {\list{}{\leftmargin=0.3in\rightmargin=0.0in}\item[]}%
  {\endlist}

\title{Constructing Code-mixed Universal Dependency Forest for\\ Unbiased Cross-lingual Relation Extraction}

\author{
Hao Fei\textsuperscript{\rm 1},  \quad
Meishan Zhang\textsuperscript{\rm 2}\Thanks{ Corresponding author},  \quad
Min Zhang\textsuperscript{\rm 2},  \quad
Tat-Seng Chua\textsuperscript{\rm 1} \\
\textsuperscript{\rm 1} Sea-NExT Joint Lab, School of Computing, National University of Singapore \\
\textsuperscript{\rm 2} Harbin Institute of Technology (Shenzhen), China \\
\tt {\{haofei37,dcscts\}@nus.edu.sg,  mason.zms@gmail.com, zhangmin2021@hit.edu.cn}
}

\begin{document}
\maketitle
\begin{abstract}
Latest efforts on cross-lingual relation extraction (XRE) aggressively leverage the language-consistent structural features from the universal dependency (UD) resource, while they may largely suffer from biased transfer (e.g., either target-biased or source-biased) due to the inevitable linguistic disparity between languages.
In this work, we investigate an unbiased UD-based XRE transfer by constructing a type of code-mixed UD forest.
We first translate the sentence of the source language to the parallel target-side language, for both of which we parse the UD tree respectively.
Then, we merge the source-/target-side UD structures as a unified code-mixed UD forest.
With such forest features, the gaps of UD-based XRE between the training and predicting phases can be effectively closed.
We conduct experiments on the ACE XRE benchmark datasets, where the results demonstrate that the proposed code-mixed UD forests help unbiased UD-based XRE transfer, with which we achieve significant XRE performance gains.
\end{abstract}

\section{Introduction}

Relation extraction (RE) aims at extracting from the plain texts the meaningful \emph{entity mentions} paired with \emph{semantic relations}.
One widely-acknowledged key bottleneck of RE is called the long-range dependence (LRD) issue, i.e., the decay of dependence clues of two mention entities with increasing distance in between \cite{culotta-sorensen-2004-dependency,zhang-etal-2018-graph,FeiGraphSynAAAI21}. 
Fortunately, prior work extensively reveals that the syntactic dependency trees help resolve LRD issue effectively, by taking advantage of the close relevance between the dependency structure and the relational RE pair \cite{miwa-bansal-2016-end,can-etal-2019-richer}.
In cross-lingual RE, likewise, the universal dependency trees \cite{MarneffeMNZ21} are leveraged as effective language-persistent features in the latest work for better transfer from source (SRC) language to target (TGT) language \cite{subburathinam-etal-2019-cross,FeiZLJ20,TaghizadehF21}.

\begin{figure}[!t]
\centering
\includegraphics[width=0.98\columnwidth]{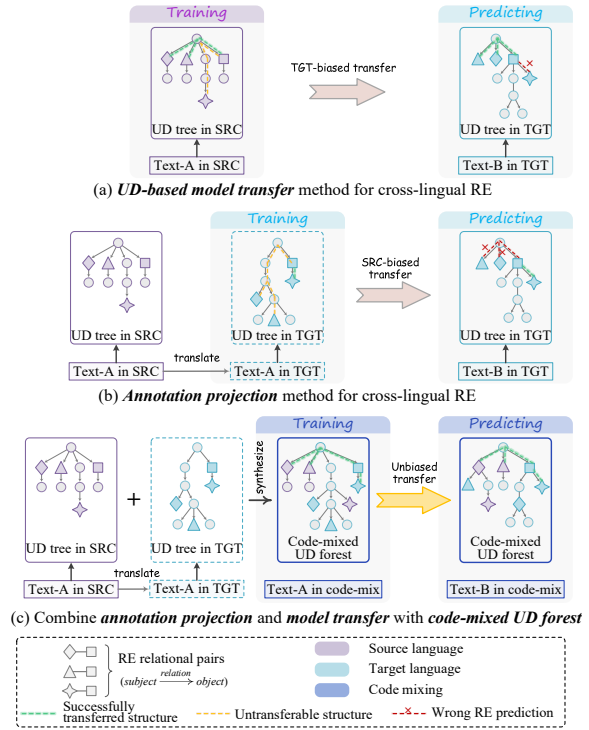}
\caption{
Model transfer fails to model the TGT-side language-specific features due to the syntactic structure discrepancy (a), while
annotation projection may overlook the SRC-side effective UD features (b).
This work combines the two methods and constructs code-mixed UD forests for unbiased cross-lingual RE (c).
}
\label{intro}
\vspace{-3mm}
\end{figure}

\begin{figure*}[!t]
\centering
\includegraphics[width=1\textwidth]{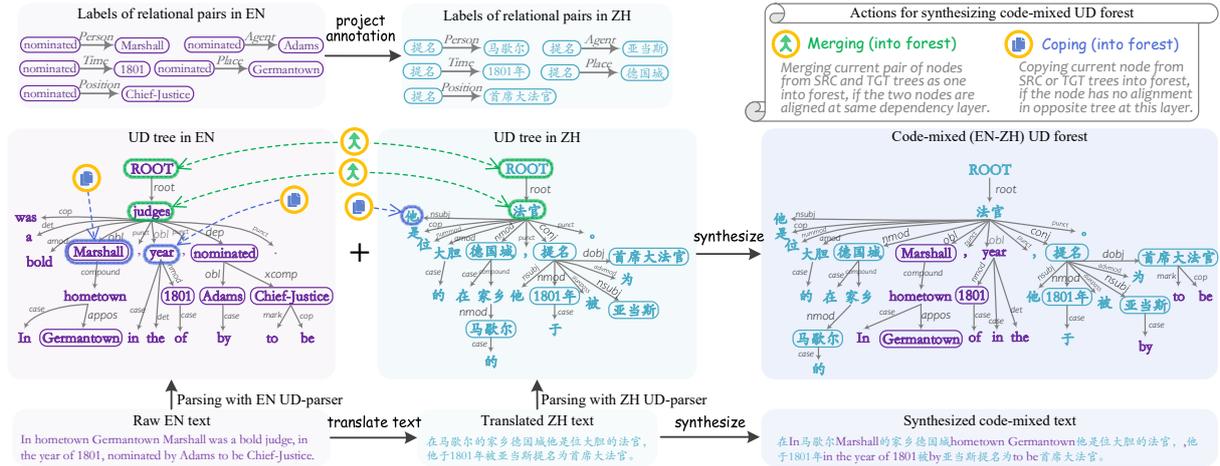}
\caption{
A real example to construct a code-mixed UD forest.
The raw sentence is selected from ACE05 data.
% for event extraction (with event type: \emph{NOMINATE}).
We exemplify the transfer from English (EN) to Chinese (ZH).
}
\label{forest}
\vspace{-4mm}
\end{figure*}

Current state-of-the-art (SoTA) XRE work leverages the UD trees based on the model transfer paradigm, i.e., training with SRC-side UD features while predicting with TGT-side UD features \cite{AhmadPC21,TaghizadehF22}.
Model transfer method transfers the shareable parts of features from SRC to TGT, while unfortunately it could fail to model the TGT-side language-specific features, and thus results in a clear \emph{TGT-side bias}.
In fact, the TGT-side bias can be exacerbated in UD-based model transfer, cf. Fig. \ref{intro}(a).
Given that UD has a universal annotation standard, inevitably, there is still a syntax discrepancy between the two languages due to their intrinsic linguistic nature.
We show (cf. $\S$\ref{Observations} for more discussion) that between the parallel sentences in English and Arabic, around 30\% words are misaligned and over 35\% UD word-pairs have no correspondence.
Such structural discrepancies consequently undermine the model transfer efficacy.

One alternative solution is using annotation projection \cite{PadoL09,kim-etal-2010-cross,mcdonald-etal-2013-universal,xiao-guo-2015-annotation}.
The main idea is directly synthesizing the pseudo TGT-side training data, so that the TGT-side linguistic features (i.e., UD trees) are well preserved.
However, it could be a double side of the sword in the annotation projection paradigm.
It manages to learn the language-specific features, while at the cost of losing some high-efficient structural knowledge from SRC-side UD, thus leading to the SRC-biased UD feature transfer.
As illustrated in Fig. \ref{intro}(b), the dependence paths in the SRC UD tree that effectively solves the LRD issues for the task are sacrificed when transforming the SRC tree into the TGT tree.

This motivates us to pursue an unbiased and holistic UD-based XRE transfer by considering both the SRC and TGT UD syntax features.
To reach the goal, in this work, we propose combining the view of model transfer and annotation projection paradigm, and constructing a type of code-mixed UD forests.
Technically, we first project the SRC training instances and TGT predicting instances into the opposite languages, respectively.
Then, we parse the parallel UD trees of both sides respectively via existing UD parsers.
Next, merge each pair of SRC and TGT UD trees together into the code-mixed UD forest, in which the well-aligned word pairs are merged to the TGT ones in the forest, and the unaligned words will all be kept in the forest.
With these code-mixed syntactic features, the gap between training and predicting phases can be closed, as depicted in Fig. \ref{intro}(c).

We encode the UD forest with the graph attention model \citep[GAT; ][]{VelickovicCCRLB18} for feature encoding.
We perform experiments on the representative XRE benchmark, ACE05 \cite{ACE5}, where the transfer results from English to Chinese and Arabic show that the proposed code-mixed forests bring significant improvement over the current best-performing UD-based system, obtaining the new SoTA results.
Further analyses verify that 1) the code-mixed UD forests help maintain the debiased cross-lingual transfer of RE task, and 2) the larger the difference between SRC and TGT languages, the bigger the boosts offered by code-mixed forests.
To our knowledge, we are the first taking the complementary advantages of annotation projection and model transfer paradigm for unbiased XRE transfer.
We verify that the gap between training and predicting of UD-based XRE can be bridged by synthesizing a type of code-mixed UD forests.
The resource can be found at \url{https://github.com/scofield7419/XLSIE/}.

% Codes and data are at \url{https://is.gd/FhRzcB}.

% \footnote{Codes and data are at \url{https://is.gd/FhRzcB}}

% \vspace{-2mm}
\section{Related Work}

\vspace{-1mm}
Different from the sequential type of information extraction (IE), e.g., named entity recognition (NER) \cite{cucerzan-yarowsky-1999-language}, RE not only detects the mentions but also recognizes the semantic relations between mentions.
RE has long received extensive research attention within the last decades \cite{zelenko-etal-2002-kernel}.
% covers a range of IE tasks, e.g., relation extraction (RE) \cite{zelenko-etal-2002-kernel}, event detection (ED) \cite{halpin-moore-2006-event} and semantic role labeling (SRL) \cite{gildea-jurafsky-2000-automatic}.
Within the community, research has revealed that the syntactic dependency trees share close correlations with RE or broad-covering information extraction tasks in structure \cite{FeiGraphSynAAAI21,Wu0RJL21,FeiLasuieNIPS22}, and thus the former is frequently leveraged as supporting features for enhancing RE.
% , i.e., for solving LRD issues.
In XRE, the key relational features between words need to be transferred between languages, which motivates the incorporation of UD tree features that have consistent annotations and principles across various languages.
Thus, UD-based systems extensively achieve the current SoTA XRE \cite{lu-etal-2020-cross,TaghizadehF21,zhang-etal-2021-benefit}.
This work inherits the prior wisdom, and leverages the UD features.

Model transfer \cite{kozhevnikov-titov-2013-cross,ni-florian-2019-neural,FeiZLJ20} and annotation projection \cite{bjorkelund-etal-2009-multilingual,mulcaire-etal-2018-polyglot,daza-frank-2019-translate,fei-etal-2020-cross,LouGYWZTX22} are two mainstream avenues in structural cross-lingual transfer track.
The former trains a model on SRC annotations and them make predictions with TGT instances, i.e., transferring the shared language-invariant features.
The latter directly synthesizes the pseudo training instances in TGT language based on some parallel sentences, in which the TGT-specific features are retained to the largest extent.
As we indicated earlier, in both two paradigms the UD tree features can be unfortunately biased during the transfer, thus leading to the underutilization of UD resource.
This work considers a holistic viewpoint, integrating both the two cross-lingual transfer schemes and combining both the SRC and TGT syntax trees by code mixing.

% In recent years' cross-lingual learning, s
Several prior studies have shown that combining the raw SRC and pseudo TGT (from projection) data for training helps better transfer.
It is shown that although the two data are semantically identical, SRC data still can offer some complementary language-biased features \cite{fei-etal-2020-cross,FeiZLJ20,zhen-etal-2021-chinese}.
Yet we emphasize that different from regular cross-lingual text classification or sequential prediction, XRE relies particularly on the syntactic structure features, e.g., UD, and thus needs a more fine-grained approach for SRC-TGT data ensembling, instead of simply instance stacking.
Thus, we propose merging the SRC and TGT syntax trees into the code-mixed forests.

Code mixing has been explored in several different NLP applications \cite{labutov-lipson-2014-generating,joshi-etal-2016-towards,banerjee-etal-2018-dataset,samanta-etal-2019-improved}, where the core idea is creating data piece containing words from different languages simultaneously.
For example, \citet{samanta-etal-2019-improved} introduce a novel data augmentation method for enhancing the recognition of code-switched sentiment analysis, where they replace the constituent phrases with code-mixed alternatives.
\citet{QinN0C20} propose generating code-switching data to augment the existing multilingual language models for better zero-shot cross-lingual tasks.
While we notice that most of the works focus on the development of code-mixed sequential texts, this work considers the one for structural syntax trees.
% (e.g., a sentence or syntax tree) 
Our work is partially similar to \citet{zhang-etal-2019-cross} on the code-mixed UD tree construction.
But ours differentiate theirs in that \citet{zhang-etal-2019-cross} target better UD parsing itself, while we aim to improve downstream tasks.

% \vspace{-1mm}
\section{Observations on UD Bias}
\label{Observations}

% \vspace{-1mm}
\subsection{Bias Source Analysis}
% \vspace{-1mm}
As mentioned, even though UD trees define consistent annotations across languages, it still falls short on wiping all syntactic bias.
This is inevitably caused by the underlying linguistic disparity deeply embedded in the language itself.
% certain specific 
% cover up
Observing the linguistic discrepancies between different languages, we can summarize them into following three levels:

% \begin{compactitem}

% \vspace{-2mm}
\paragraph{1) Word-level Changes.}
\begin{compactitem}
    \item \textbf{Word number.} The words referring to same semantics in different languages vary, e.g., in English one single-token word may be translated in Chinese with more than one token. 
    \item \textbf{Part of speech.} In different languages a parallel lexicon may come with different part of speech.
    \item \textbf{Word order.} Also it is a common case that the word order varies among parallel sentences in different languages.
\end{compactitem}

% \vspace{-2mm}
\paragraph{2) Phrase-level Change.}
\begin{compactitem}
    \item \textbf{Modification type.} A modifier of a phrasal constituent can be changed when translating into another languages. For example, in English, `in the distance' is often an adverbial modifier, while its counterpart in Chinese \begin{CJK*}{UTF8}{gbsn}`遥远的'\end{CJK*} plays a role of an attribute modifier.

    \item \textbf{Change of pronouns.}  English grammar has strict structure, while in some other languages the grammar structures may not strict. 
    For example, in English, it is often case to use relative pronouns (e.g., which, that, who) to refer to the prior mentions, while in other languages, such as Chinese, the personal pronouns (e.g., which, that, who) will be used to refer the prior mentions.
    
    \item \textbf{Constituency order change.}  Some constituent phrases will be reorganized and reordered from one language to another language, due to the differences in grammar rules.
\end{compactitem}

% \vspace{-2mm}
\paragraph{3) Sentence-level Change.}
\begin{compactitem}
    \item \textbf{Transformation between active and passive sentences.}  In English it could be frequent to use the passive forms of sentences, while being translated into other languages the forms will be transformed into active types, where the words and phrases in the whole sentences can be reversed.
    
    \item \textbf{Transformation between clause and main sentence.} In English the attributive clauses and noun clauses are often used as subordinate components, while they can be translated into two parallel clauses in other languages.
    
    \item \textbf{Change of reading order of sentences.} The majority of the languages in this world have the reading order of from-left-to-right, such as English, French, etc. But some languages, e.g., under Afro-Asiatic family, Arabic, Hebrew, Persian, Sindhi and Urdu languages read from right to left.
\end{compactitem}

% \end{compactitem}

\begin{figure}[!t]
\centering
\includegraphics[width=1\columnwidth]{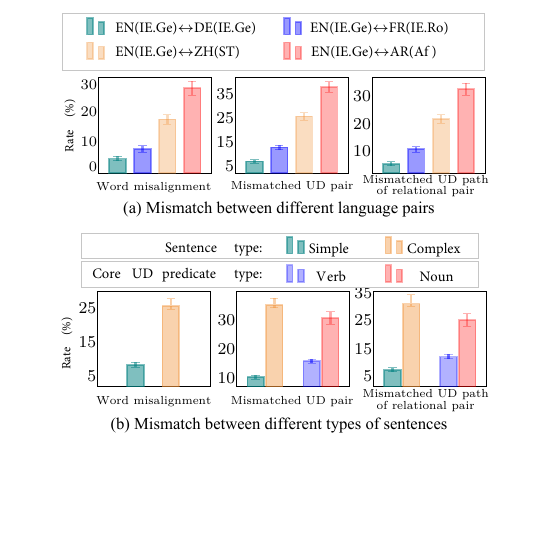}
% \vspace{-2mm}
\caption{
Statistics of mismatching items of UD trees.
}
\vspace{-3mm}
\label{Mismatch-stat-1}
\end{figure}

\subsection{UD Bias Statistics}
In Fig. \ref{Mismatch-stat-1} we present the statistics of such bias between the parallel UD trees in different languages, such as the misaligned words, mismatched UD ($w_i^{\curvearrowright}w_j$) pair and UD path of ($e_s^{\curvearrowright}\cdots^{\curvearrowright}e_o$) relational pair.
Fig. \ref{Mismatch-stat-1}(a) reveals that languages under different families show distinct divergences.
And the more different of languages, the greater the divergences (e.g., English to Arabic).
Fig. \ref{Mismatch-stat-1}(b) indicates that complex sentences (e.g., compound sentences) bring larger bias; and in the real world, complex sentences are much more ubiquitous than simple ones.
Also, the mismatch goes worse when the UD core predicates are nouns instead of verbs.

\begin{algorithm*}[!t]
\caption{Process of constructing code-mixed UD forests}
\label{process-structure-induction}
\begin{algorithmic}[1]
\REQUIRE ~~
$T^{\texttt{SRC}}$, $T^{\texttt{TGT}}$, $M$, threshold $\theta$, empty forest $\mathcal{F}=\Phi$.
\ENSURE ~~
Code-mixed UD forest $\mathcal{F}$. \\
% \vspace{3pt}\rule[10pt]{13cm}{0.6pt}\vspace{-1em} 
\vspace{3pt}\rule[10pt]{14.5cm}{0.6pt}\vspace{-9pt} 

\STATE \textbf{def} \textbf{Construct } ($\mathcal{T}^{\texttt{SRC}}$, $\mathcal{T}^{\texttt{TGT}}$, $M$, $\mathcal{F}$)  \hfill  \textcolor{mypurplex}{$\triangleright$ breadth-first top-down traverse.} \
% \bindent
\begin{ALC@g}
    \STATE is\_root = True \hfill  \textcolor{mypurplex}{$\triangleright$ a flag for traversing the predicate only once.}
    \STATE $\mathcal{F}.w_{cur}$ = ROOT \hfill  \textcolor{mypurplex}{$\triangleright$  creating ROOT node for $\mathcal{F}$.}
    % $\mathcal{T}^C.V$.Add(\{$w_1,\cdots,w_n$\});  \hfill \textcolor{mypurplex}{$\triangleright$ Adding word leave nodes.} \
    \STATE opt\_nodes = Queue.Init() \hfill  \textcolor{mypurplex}{$\triangleright$  creating a queue for breadth-first search.}
    \WHILE{($\mathcal{T}^{\texttt{SRC}} \ne \Phi$) or ($\mathcal{T}^{\texttt{TGT}} \ne \Phi$) or (opt\_nodes$\ne \Phi$)} 
        \IF{is\_root}
            \STATE $w_{merged}$ = Merge($\mathcal{T}^{\texttt{SRC}}$.ROOT, $\mathcal{T}^{\texttt{TGT}}$.ROOT) \hfill  \textcolor{mypurplex}{$\triangleright$  merging from ROOT in $\mathcal{T}^{\texttt{SRC}}$ and $\mathcal{T}^{\texttt{TGT}}$.}
            % two core predicates
            \STATE $w_{merged}$.next$^{\texttt{SRC}}$ = $\mathcal{T}^{\texttt{SRC}}$.ROOT.GetChildNodes()
            \STATE $w_{merged}$.next$^{\texttt{TGT}}$ = $\mathcal{T}^{\texttt{TGT}}$.ROOT.GetChildNodes()
            \STATE $\mathcal{F}.w_{cur}$.SetChild($w_{merged}$, `root')
            \STATE opt\_nodes.enqueue($w_{merged}$)
            \STATE is\_root = False
        \ELSE
            \STATE $\mathcal{F}.w_{cur}$ = opt\_nodes.dequeue()
            \STATE aligned\_pairs, nonaligned\_nodes = AlignSearch($\mathcal{F}.w_{cur}$.next$^{\texttt{SRC}}$, $\mathcal{F}.w_{cur}$.next$^{\texttt{TGT}}$, $M$)
            \FOR{( $w^{\texttt{SRC}}_i$, $w^{\texttt{TGT}}_j$, $arc$ ) $\in$ aligned\_pairs}
                \STATE $w_{merged}$ = Merge($w^{\texttt{SRC}}_i$, $w^{\texttt{TGT}}_j$)
                \STATE $w_{merged}$.next$^{\texttt{SRC}}$ = $w^{\texttt{SRC}}_i$.GetChildNodes()
                \STATE $w_{merged}$.next$^{\texttt{TGT}}$ = $w^{\texttt{TGT}}_j$.GetChildNodes()
                \STATE $\mathcal{F}.w_{cur}$.SetChild($w_{merged}$, $arc$)
                \STATE opt\_nodes.enqueue($w_{merged}$)
            \ENDFOR
            \FOR{$w_i$ $\in$ nonaligned\_nodes}
                \STATE $\mathcal{F}.w_{cur}$.SetChild($w_i$, $w_i$.$arc$) \hfill  \textcolor{mypurplex}{$\triangleright$  action `\emph{Coping into forest}' for non-aligned words.}%: copying all the non-aligned nodes into forest. }
            \ENDFOR
            
        \ENDIF
        % \ENDFOR
    \ENDWHILE
    \STATE \textbf{return} $\mathcal{F}$ \
% \eindent
\end{ALC@g}

\vspace{7pt}

\STATE \textbf{def} \textbf{Merge} ($w^{\texttt{SRC}}_a$, $w^{\texttt{TGT}}_b$)   \hfill  \textcolor{mypurplex}{$\triangleright$  action `\emph{Merging into forest}' for aligned words.} \
\begin{ALC@g}
    \STATE \textbf{return} $w^{\texttt{TGT}}_b$ \hfill \textcolor{mypurplex}{$\triangleright$ for two aligned word, returning the TGT-side word.}\
\end{ALC@g}

\vspace{7pt}

\STATE \textbf{def} \textbf{AlignSearch} (nodes\_a, nodes\_b, $M$)  \hfill \textcolor{mypurplex}{$\triangleright$ preparing the aligned word pairs in $\mathcal{T}^{\texttt{SRC}}$ and $\mathcal{T}^{\texttt{TGT}}$.} \
\begin{ALC@g}
    \STATE aligned\_pairs = [] \
    \FOR{ $m_{i\leftrightarrow j} \in M$}
        \IF{ $m_{i\leftrightarrow j} > \theta$}% \& (nodes\_a[i].$arc$ == nodes\_a[j].$arc$)}
            \STATE aligned\_pairs.Append(nodes\_a[i], nodes\_b[j], nodes\_b[i].$arc$ ) \
            \STATE nodes\_a.Remove($w_i$) \
            \STATE nodes\_a.Remove($w_j$) \
        \ENDIF
    \ENDFOR
    \STATE nonaligned\_nodes = nodes\_a.union(nodes\_b)  \hfill \textcolor{mypurplex}{$\triangleright$ words with no salient alignments.} \
    \STATE \textbf{return} aligned\_pairs, nonaligned\_nodes \
\end{ALC@g}
\end{algorithmic}
\end{algorithm*}

% \vspace{-1mm}
\section{Code-mixed UD Forest Construction}
\label{Code-mixed UD Forest Construction}

To eliminate such discrepancies for unbiased UD-feature transfer, we build the code-mixed UD forests, via the following six steps.

\vspace{3pt}
\textbf{$\blacktriangleright$ Step 1: translating a sentence ${x^{\texttt{\scriptsize Src}}}$ in SRC language to the one ${x^{\overline{\texttt{\scriptsize Tgt}}}}$ in TGT language.}\footnote{Vice versa for the direction from TGT to SRC language.}
This step is to generate a pseudo parallel sentence pair in both TGT and SRC languages.
We accomplish this by using the state-of-the-art \emph{Google Translation API}.\footnote{\url{https://translate.google.com}, Sep. 10 2022}
We denote the parallel sentences as <${x^{\texttt{\scriptsize Src}}}$,${x^{\overline{\texttt{\scriptsize Tgt}}}}$> or <${x^{\overline{\texttt{\scriptsize Src}}}}$,${x^{\texttt{\scriptsize Tgt}}}$>.

\vspace{3pt}
\textbf{$\blacktriangleright$ Step 2: obtaining the word alignment scores.}
Meanwhile, we employ the Awesome-align toolkit\footnote{\url{https://github.com/neulab/awesome-align}} to obtain the word alignment confidence $M$=$\{m_{i\leftrightarrow j}\}$ between word pair $w_i \in x^{\texttt{\scriptsize Src}}$ and $w_j \in x^{\overline{\texttt{\scriptsize Tgt}}}$ in parallel sentences.

\vspace{3pt}
\textbf{$\blacktriangleright$ Step 3: parsing UD trees for parallel sentences.}
Then, we use the UD parsers in SRC and SRC languages respectively to parse the UD syntax trees for two parallel sentences, respectively.
We adopt the UDPipe\footnote{\url{https://github.com/bnosac/udpipe},
Universal Dependencies 2.3 models: english-ewtud-2.3-181115.udpipe, chinese-gsd-ud-2.3-181115.udpipe, arabic-padt-ud-2.3-181115.udpipe.
} as our UD parsers, which are trained separately on different UD annotated data\footnote{\url{https://universaldependencies.org/}}.
We denote the SRC UD tree as $\mathcal{T}^{\texttt{\scriptsize Src}}$, and the pseudo TGT UD tree as $\mathcal{T}^{\overline{\texttt{\scriptsize Tgt}}}$.
Note that the UD trees in all languages share the same dependency labels, i.e., with the same (as much as possible) annotation standards.
In Appendix \S\ref{The universal dependency labels} we list the dependency labels which are the commonly occurred types.
% Please refer to Stanford dependency\footnote{\url{https://nlp.stanford.edu/software/dependencies_manual.pdf}} for more details about the dependency labels.

\vspace{3pt}
\textbf{$\blacktriangleright$ Step 4: projecting and merging the labels of training data.}
For the training set, we also need to project the annotations (relational subject-object pairs) of sentences in SRC languages to TGT pseudo sentences.
Note that this step is not needed for the testing set.
The projection is based on the open source\footnote{\url{https://github.com/scofield7419/XSRL-ACL}}, during which the word alignment scores at step-2 are used.
We can denote the SRC annotation as $y$, and the pseudo TGT label as $\overline{y}$.
We then merge the annotation from both SRC and TGT viewpoints, into the code-mixed one $Y$, for later training use.
Specifically, for the node that is kept in the final code-mixed forest, we will keep its labels; and for those nodes that are filtered, the annotations are replaced by their correspondences.
% Note that the annotation merge is quite straightforward.

\vspace{1pt}
\textbf{$\blacktriangleright$ Step 5: merging the SRC and TGT UD trees into a code-mixed forest.}
Finally, based on the SRC UD tree and the TGT UD tree, we construct the code-mixed UD forest.
We mainly perform breadth-first top-down traversal over each pair of nodes $\mathcal{T}^{\texttt{\scriptsize Src}}$ and $\mathcal{T}^{\overline{\texttt{\scriptsize Tgt}}}$, layer by layer.
The traversal starts from their \emph{ROOT} node.
We first create a \emph{ROOT} node as the initiation of the code-mixed forest.
We design two types of actions for the forest merging process:

\vspace{-3mm}
\begin{myquote}
    $\bullet$ \textbf{\texttt{Merging}} current pair of nodes $w_i \in \mathcal{T}^{\texttt{\scriptsize Src}}$ from SRC tree and $w_j \in \mathcal{T}^{\overline{\texttt{\scriptsize Tgt}}}$ from TGT tree into the forest $\mathcal{F}$, if the current two nodes are confidently aligned at same dependency layer.
    % To decide if two nodes are highly aligned, w
    We check the word alignment confidence $m_{i\leftrightarrow j}$ between the two nodes, and if the confidence is above a pre-defined threshold $\theta$, i.e., $m_{i\leftrightarrow j}>\theta$, we treat them as confidently aligned.
\end{myquote}
\vspace{-3mm}

\vspace{-3mm}
\begin{myquote}
    $\bullet$ \textbf{\texttt{Copying}} current node from SRC tree $\mathcal{T}^{\texttt{\scriptsize Src}}$ or TGT tree $\mathcal{T}^{\overline{\texttt{\scriptsize Tgt}}}$ into the forest $\mathcal{F}$, once the node has no significant alignment in the opposite tree at this layer.
\end{myquote}
\vspace{-3mm}

In Algorithm \ref{process-structure-induction} we formulate in detail the process of code-mixed forest construction.
Also, we note that when moving the nodes from two separate UD trees into the forest, the attached dependency labels are also copied.
When two nodes are merged, we only choose the label of the TGT-side node.
Finally, the resulting forest $\mathcal{F}$ looks like code-mixing, and is structurally compact.

\vspace{3pt}
\textbf{$\blacktriangleright$ Step 6: assembling code-mixed texts.}
Also we need to synthesize a code-mixed text $X$ based on the raw SRC text ${x^{\texttt{\scriptsize Src}}}$ and the pseudo TGT text ${x^{\overline{\texttt{\scriptsize Tgt}}}}$.
The code-mixed text $X$ will also be used as inputs together with the forest, into the forest encoder.
We directly replace the SRC words with the TGT words that have been determined significantly aligned at Step-5.

% \vspace{-1mm}
\section{XRE with Code-mixed UD Forest}
\label{XRE with Code-mixed UD Forest}

% \vspace{-4pt}
% \paragraph{Problem Formulation}
Along with the UD forest $\mathcal{F}^{\texttt{\scriptsize Src}}$, we also assemble the code-mixed sequential text $X^{\texttt{\scriptsize Src}}$ from the SRC and translated pseudo-TGT sentences (i.e., ${x^{\texttt{\scriptsize Src}}}$ and ${x^{\overline{\texttt{\scriptsize Tgt}}}}$), and the same for the TGT sentences $X^{\texttt{\scriptsize Tgt}}$.
An XRE system, being trained with SRC-side annotated data (<$X^{\texttt{\scriptsize Src}}$, $\mathcal{F}^{\texttt{\scriptsize Src}}$>, $Y^{\texttt{\scriptsize Src}}$), needs to determine the label $Y^{\texttt{\scriptsize Tgt}}$ of relational pair $e_s^{\overset{r}{\curvearrowright}}e_o$ given a TGT sentence and UD forest (<$X^{\texttt{\scriptsize Tgt}}$, $\mathcal{F}^{\texttt{\scriptsize Tgt}}$>).

The XRE system takes as input $X$=$\{w_i\}_n$ and $\mathcal{F}$.
We use the multilingual language model (MLM) for representing the input code-mixed sentence $X$:
\begin{equation}\small
\bm{H} = \{ \bm{h}_1,\cdots, \bm{h}_n \} = \text{MLM} (X)  \,,
\end{equation}
where $X$ is the code-mixed sentential text.
We then formulate the code-mixed forest $\mathcal{F}$ as a graph, $G$=<$E,V$>, where $E$=$\{e_{i,j}\}_{n\times n}$ is the edge between word pair (i.e., initiated with $e_{i,j}$=0/1, meaning dis-/connecting), $V$=$\{w_i\}_n$ are the words.
We main the node embeddings $\bm{r}_i$ for each node $v_i$.
We adopt the GAT model \cite{VelickovicCCRLB18} for the backbone forest encoding:
\setlength\abovedisplayskip{3pt}
\setlength\belowdisplayskip{3pt}
\begin{equation}\small\label{GAT}
\setlength\abovedisplayskip{3pt}
\setlength\belowdisplayskip{3pt}
 \rho_{i,j} = \text{Softmax}( \text{GeLU}( \bm{U}^T [\bm{W}_{1} \bm{r}_{i} ; \bm{W}_{2} \bm{r}_{j} ]) ) \,,
\end{equation}
\begin{equation}\small
\setlength\abovedisplayskip{3pt}
\setlength\belowdisplayskip{3pt}
 \bm{u}_{i} = \sigma( \sum_{j} \rho_{i,j}  \bm{W}_3 \bm{r}^1_{j})  \,, 
\end{equation}
where $\bm{W}_{3/4/5}$ and $\bm{U}$ are all trainable parameters.
$\sigma$ is the sigmoid function.
GeLU is a Gaussian error linear activation function.
Note that the first-layer representations of $\bm{r}_i$ is initialized with $\bm{h}_i$.
$\bm{H}$ and $\bm{U}$ are then concatenated as the resulting feature representation:
\begin{equation}\small
\setlength\abovedisplayskip{3pt}
\setlength\belowdisplayskip{3pt}
 \bm{\hat{H}} = \bm{H}\oplus\bm{U} \,.
\end{equation}

XRE aims to determine the semantic relation labels between two given mention entities.
For example, given a sentence `\emph{John Smith works at Google}', RE should identify that there is a relationship of "works at" between the entities "John Smith" and "Google".
Our XRE model needs to predict the relation label $y$.
We adopt the biaffine decoder \cite{DozatM17} to make prediction:
\begin{equation}\small\label{biaffine}
\setlength\abovedisplayskip{2pt}
\setlength\belowdisplayskip{2pt}
 y = \text{Softmax}( \bm{h}^T_s \cdot \bm{W}_1 \cdot \bm{h}_o + \bm{W}_2 \cdot \text{Pool}(\bm{\hat{H}})) \,.
\end{equation}
Here both $\bm{h}_s$ and $\bm{h}_o$ are given.

\begin{table}[!t]
\fontsize{9.5}{11}\selectfont
 \setlength{\tabcolsep}{3.mm}

\fontsize{9.5}{11.5}\selectfont
 \setlength{\tabcolsep}{3.5mm}
\begin{center}
% \resizebox{1\columnwidth}{!}{
  \begin{tabular}{lccc}
\toprule
\bf Language  &\bf Train & \bf Dev & \bf Test \\
\midrule
 EN  & 479  & 60 & 60 \\
 ZH & 507   & 63 & 63 \\
 AR & 323  &  40 & 40 \\
\bottomrule
\end{tabular}
% }
\end{center}
  \caption{
Data statistics.
The numbers are documents.
 }
  \label{data-stats-re}
\end{table}

% \vspace{-1mm}
\section{Experiments}
\label{Experiments}

\begin{table*}[!t]
\fontsize{10}{12}\selectfont
 \setlength{\tabcolsep}{4mm}
\begin{center}
  \begin{tabular}{llccccl}
\toprule
 & & \textcolor{nmtextpurple}{\bf SRC} & \textcolor{nmtextblue}{\bf TGT} & \textbf{EN}$\to$\textbf{ZH} & \textbf{EN}$\to$\textbf{AR} & \textbf{AVG} \\
\midrule

\multicolumn{6}{l}{\textbf{$\blacktriangleright$ Model Transfer}} \\

\quad M1 &	TxtBaseline &	\checkmark	 & &	55.8 &	63.8 & 59.8 \\
\quad M2 & 	SynBaseline(+$\mathcal{T}$) & 	\checkmark  & & 59.2 & 	65.2 & 	62.2 \scriptsize{(+2.4)} \\
\quad M3 & SoTA XRE  &	\checkmark &	& 	58.0 & 	66.8 &	62.4 \\
\cdashline{1-7}

\multicolumn{6}{l}{\textbf{$\blacktriangleright$ Annotation Projection}} \\

\quad M4&	TxtBaseline&	&\checkmark&	58.3&	66.2& 62.3 \\
\quad M5&	SynBaseline(+$\mathcal{T}$)&	&\checkmark &61.4&	67.4&	64.4 \scriptsize{(+2.1)} \\
\cdashline{1-7}

\multicolumn{6}{l}{\textbf{$\blacktriangleright$ Model Transfer + Annotation Projection}} \\

\quad M6&	SynBaseline(+$\mathcal{T}$)&	\checkmark&	\checkmark&	57.8&	64.0&	60.9 \\
\rowcolor{nmgray} \quad M7 (Ours) &	SynBaseline(+$\mathcal{F}$) &	\checkmark&	\checkmark&	63.7&	70.7&	67.2 \scriptsize{(+6.3)} \\
% \cdashline{1-18}
\quad M8& \quad	w/o code-mixed text&	\checkmark&	\checkmark& 61.6&	68.2&	64.9 \scriptsize{(-2.3)}\\

\bottomrule
\end{tabular}
\end{center}
\vspace{-1mm}
  \caption{
Main results of cross-lingual RE transfer tasks from English language to other languages, by different models and features.
M6 uses two separate instances (texts and UD trees) for training, including the raw SRC one and the pseudo TGT one.
M7 uses the SRC-TGT merged one as ours, i.e., code-mixed texts and forests.
 }
  \label{main-1}
  % \vspace{-3mm}
\end{table*}

\vspace{-1mm}
\subsection{Setups}
\label{Setups}

\vspace{-2pt}
We consider the ACE05 \cite{ACE5} dataset, which includes English (EN), Chinese (ZH) and Arabic (AR).
We give the data statistics in Table \ref{data-stats-re}
The multilingual BERT is used.\footnote{\url{https://huggingface.co}, base, cased version}
We use two-layer GAT for forest encoding, with a 768-d hidden size.
We mainly consider the transfer from EN to one other language.
Following most cross-lingual works \cite{FeiZLJ20,AhmadPC21}, we train the XRE model with fixed 300 iterations without early-stopping.
We make comparisons between three setups: 
1) using only raw SRC training data with the model transfer, 
2) using only the pseudo TGT (via annotation projection) for training, 
and 3) using both the above SRC and TGT data.
Each setting uses both the texts and UD tree (or forest) features.
The baseline uses the same GAT model for syntax encoding, marked as \emph{SynBaseline}.
For setup 1)\&2) we also test the transfer with only text inputs, removing the syntax features, marked as \emph{TxtBaseline}.
Besides, for setup 1) we cite current SoTA performances as references.
We use F1 to measure the RE performance, following \citet{AhmadPC21}.
All experiments are undergone five times and the average value is reported.

\begin{table}[!t]
\begin{center}
\fontsize{9.5}{11}\selectfont
 \setlength{\tabcolsep}{5mm}
% \resizebox{1\columnwidth}{!}{
  \begin{tabular}{cccc}
\toprule
% \midrule
 &\bf EN &\bf ZH & \bf  AR  \\
\midrule
\multicolumn{3}{l}{$\bullet$\bf Sequential Distance} \\
 &4.8 & 3.9  & 25.8  \\
 \hline
\multicolumn{3}{l}{$\bullet$\bf Syntactic Distance} \\
 & 2.2 & 2.6 & 5.1  \\
\hline
\end{tabular}
% }
\end{center}
  \caption{
Sequential and syntactic (shortest dependency path) distances (words) between the subjects and objects of the relational triplets.
 }
\vspace{-3mm}
  \label{distances-stats}
\end{table}

% \vspace{-1mm}
\subsection{Data Inspection}

We also show in Table \ref{distances-stats} the differences in average sequential and syntactic (shortest dependency path) distances between the subjects and objects of the relational triplets.
As seen, the syntactic distances between subject-object pairs are clearly shortened in the view of syntactic dependency trees, which indicates the imperative to incorporate the tree structure features.
However, the syntactic distances between different languages vary, i.e., more complex languages have longer syntactic distances.
Such discrepancy reflects the necessity of employing our proposed UD debiasing methods to bridge the gap.

\begin{figure}[!t]
\centering
% \vspace{-2mm}
\includegraphics[width=0.98\columnwidth]{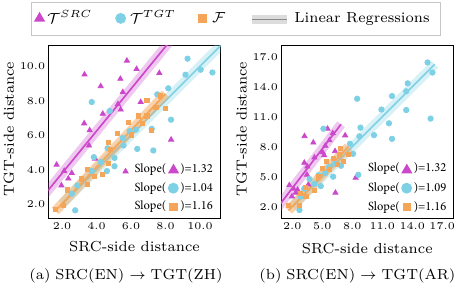}
% \vspace{-3mm}
\caption{
Change of syntax distance (shortest path) of relational pair in different UD trees.
}
% \vspace{-6mm}
\label{correlation}
\end{figure}

\begin{table*}[!t]
\fontsize{9.5}{12}\selectfont
\setlength{\tabcolsep}{4.5mm}
\begin{center}
% \resizebox{1\textwidth}{!}{
\begin{tabular}{lccccc}
% \hline
\toprule
& \multicolumn{5}{c}{\bf Words per Sentence} \\
\cmidrule(r){2-6}
& \multicolumn{3}{c}{\bf Before Merging}  & \multicolumn{2}{c}{\bf After Merging} \\
\cmidrule(r){2-4}\cmidrule(r){5-6}
& \bf SRC (EN) & \bf TGT &\bf Sum & 	\bf Code-mixed & \bf Merged (Rate) \\
\midrule
 EN-ZH & 15.32 & 24.91 	& 40.23 & 31.63	& 8.6 (21.4\%) \\
 EN-AR & 15.32 & 33.12	 	& 48.44	& 40.44	&  8.0 (16.6\%) \\

\bottomrule
  \end{tabular}
% }
\end{center}
\vspace{-5pt}
\caption{
The statistics of the words before and after constructing code-mixed data.
}
\vspace{-5pt}
  \label{forest-Statistics1}
\end{table*}

% \vspace{-5mm}
\subsection{Main Results}

% \vspace{-4pt}
From Table \ref{main-1}, we can see that UD features offer exceptional boosts (M1 vs. M2, M4 vs. M5).
And annotation projection methods outperform model transfer ones (i.e., M1\&M2\&M3 vs. M4\&M5) by offering direct TGT-side features.
Interestingly, in both two transfer paradigms, the improvements from UD become weak on the language pairs with bigger divergences.
For example, the improvement on EN$\to$DE outweighs the ones on EN$\to$ZH.
Furthermore, using our proposed code-mixed syntax forests is significantly better than using standalone SRC or TGT (or the simple combination) UD features (M7 vs. M2\&M5\&M6) on all transfers with big margins.
For example, our system outperforms SoTA UD-based systems with averaged +4.8\%(=67.2-62.4) F1.
This evidently verifies the necessity to create the code-mixed forests, i.e., bringing unbiased UD features for transfer.
Also, we find that the more the difference between the two languages, the bigger the improvements from forests.
The ablation of code-mixed texts also shows the contribution of the sequential textual features, which indirectly demonstrates the larger efficacy of the structural code-mixed UD forests.

% \vspace{-2.5mm}
\subsection{Probing Unbiasedness of Code-mixed UD Forest}
Fig. \ref{correlation} plots the change of the syntax distances of RE pairs during the transfer with different syntax trees.
We see that the use of SRC UD trees shows clear bias (with larger inclination angles) during the transfer, while the use of TGT UD trees and code-mixed forests comes with less change of syntax distances.
Also, we can see from the figure that the inference paths between objects and subjects of RE tasks are clearly shortened with the forests (in orange color), compared to the uses of SRC/TGT UD trees.

\subsection{Change during Code-mixed UD Forest Merge}

Here we make statistics of how many words are merged and kept during the UD tree merging, respectively.
The statistics are shown in Table \ref{forest-Statistics1}.
We can see that the distance between EN-ZH is shorter than that between EN-AR.
For example, the length of code-mixed EN-ZH UD forests (sentences) is 31.63, while for EN-AR the length is 40.44.
Also, EN-ZH UD forests have a higher to 21.4\% merging rate, while EN-AR UD forests have 16.6\% merging rate.
This demonstrates that the more divergences of languages, the lower the merging rate of the code-mixed forest.

\subsection{Impacts of $\theta$ on Controlling the Quality of Merged Forest}

\begin{figure}[!t]
\centering
\includegraphics[width=1.0\columnwidth]{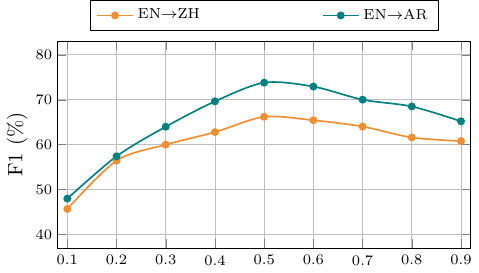}
\caption{
Transfer performances by using code-mixed forests generated with different merging rates ($\theta$).
}
\vspace{-5pt}
\label{threshold-theta-re}
\end{figure}

In $\S$\ref{Code-mixed UD Forest Construction} of step-5, we describe that we use a threshold $\theta$ to control the aligning during the UD tree merging.
Intuitively, the large the threshold $\theta$, the lower the alignment rate.
% , and thus the al
When $\theta \to 0$, most of the SRC and TGT nodes in two parallel UD trees can find their counterparts but the alignments are most likely to be wrong, thus hurting the quality of the resulting code-mixed UD forests.
When $\theta \to 1$, none of the SRC and TGT nodes in two parallel UD trees can be aligned, and both two UD trees are copied and co-existed in the resulting code-mixed UD forests.
In such case, the integration of such forests is equivalent to the annotation projection methods where we directly use both the raw SRC UD feature and the translated pseudo TGT UD tree feature.
% , and the results is as shown in Table \ref{main-1} with M6.
In Fig. \ref{threshold-theta-re} we now study the influences of using different code-mixed forest features generated with different merging rates ($\theta$). 
We see that with a threshold of $\theta$=0.5, the performances are consistently the best.

\subsection{Performances on Different Types of Sentence}

In Table \ref{sentence-type-re} we show the results under different types of sentences.
We directly select 500 short sentences (with length $<$ 12) as simple sentences;
and select 500 lengthy sentences (with length $>$ 35) as complex sentences.
As can be seen, with the code-mixed forest features, the system shows very notable improvements in complex sentences.
For example, on the EN$\to$ZH we obtain 15.9(=57.2-41.3)\% F1 improvement,
and on the EN$\to$AR the boost increases strikingly to 25.2(=67.3-42.1)\% F1.
However, such enhancements are not very significant in handling simple sentences.
This indicates that the code-mixed UD forest features can especially enhance the effectiveness on the hard case, i.e., the transfer between those pairs with greater divergences will receive stronger enhancements from our methods.

\begin{table}[!t]
\begin{center}
\fontsize{9}{12}\selectfont
 \setlength{\tabcolsep}{3.5mm}
% \resizebox{1\columnwidth}{!}{
  \begin{tabular}{lcc}
\toprule
\bf  &  \bf EN$\to$ZH &\bf EN$\to$AR   \\
\midrule
\multicolumn{3}{l}{$\bullet$ \bf Simple Sentence } \\
\quad SynBaseline(+$\mathcal{T}^{\scriptsize{\textcolor{nmtextpurple}{SRC}}}$) &  	66.1 &	78.2   \\
\quad SynBaseline(+$\mathcal{T}^{\scriptsize{\textcolor{nmtextblue}{TGT}}}$) &		68.7&	80.6 \\
\quad SynBaseline(+$\mathcal{F}$)&	71.3&	82.4 \\

\hline
\multicolumn{3}{l}{$\bullet$ \bf Complex Sentence } \\
\quad SynBaseline(+$\mathcal{T}^{\scriptsize{\textcolor{nmtextpurple}{SRC}}}$) &  		39.5 &	37.4  \\
\quad SynBaseline(+$\mathcal{T}^{\scriptsize{\textcolor{nmtextblue}{TGT}}}$) &			41.3&	42.1 \\
\quad SynBaseline(+$\mathcal{F}$)&		57.2&	67.3 \\

\bottomrule
\end{tabular}
% }
\end{center}
\vspace{-5pt}
\caption{
Comparisons under different types of sentences.
}
\vspace{-5pt}
\label{sentence-type-re}
\end{table}

% \vspace{-6pt}
\section{Conclusion and Future Work}

% \vspace{-8pt}
Universal dependencies (UD) have been served as effective language-consistent syntactic features for cross-lingual relation extraction (XRE).
In this work, we reveal the intrinsic language discrepancies with respect to the UD structural annotations, which limit the utility of the UD features.
We enhance the efficacy of UD features for an unbiased UD-based transfer, by constructing code-mixed UD forests from both the source and target UD trees.
Experimental results demonstrate that the UD forests effectively debias the syntactic disparity in the UD-based XRE transfer, especially for those language pairs with larger gaps.

Leveraging the syntactic dependency features is a long-standing practice for strengthening the performance of RE tasks.
In this work, we propose a novel type of syntactic feature, code-mixed UD forests, for cross-lingual relation extraction.
We note that this feature can be applied broadly to other cross-lingual structured information extraction tasks that share the same task definition besides RE, such as event detection (ED) \cite{halpin-moore-2006-event} and semantic role labeling (SRL) \cite{gildea-jurafsky-2000-automatic}.
Besides, how to further increase the utility of the UD forests with a better modeling method is a promising research direction, i.e., filtering the noisy structures in the UD forests.

\vspace{-1mm}
\section*{Acknowledgments}

\vspace{-2mm}
This research is supported by the National Natural Science Foundation of China (No. 62176180), and also the Sea-NExT Joint Lab.

\vspace{-1mm}
\section*{Limitations}

\vspace{-2mm}
Although showing great prominence, our proposed method has the following limitations.
First of all, our method relies on the availability of annotated UD trees of both the source and target languages, as we need to use the annotations to parse the syntax trees for our own sentences.
Fortunately, UD project covers over 100 languages, where most of the languages, even the minor ones, will have the UD resources. 
At the same time, our method will be influenced by the quality of UD parsers.
Secondly, our method also uses the external translation systems to produce the pseudo parallel sentences, where our method may largely subject to the quality of the translators.
Again luckily, current neural machine translation systems have been well developed and established, i.e., Google Translation.
Only when handling very scare languages where the current translation systems fail to give satisfactory performances, our method will fail.

\vspace{-1mm}
\section*{Ethics Statement}

\vspace{-2mm}
In this work, we construct a type of code-mixed UD forest based on the existing UD resources.
We note that all the data construction has been accomplished automatically, and we have not created any new annotations with additional human labor.
Specifically, we use the UD v2.10 resource, which is a collection of linguistic data and tools that are open-sourced.
Each of treebanks of UD has its own license terms, including the \emph{CC BY-SA 4.0}\footnote{\url{	http://creativecommons.org/licenses/by-sa/4.0/}} and \emph{CC BY-NC-SA 2.5-4.0}\footnote{\url{	http://creativecommons.org/licenses/by-nc-sa/4.0/}} as well as \emph{GNU GPL 3.0}\footnote{\url{http://opensource.org/licenses/GPL-3.0}}.
Our use of UD treebanks comply with all these license terms is at non-commercial purpose.
The software tools (i.e., UDPipe parsers) are provided under \emph{GNU GPL V2}.
Our use of UDPipe tools complies with the term.

\vspace{-3mm}

\bibliographystyle{acl_natbib}
\bibliography{ref}

\begin{thebibliography}{40}
\expandafter\ifx\csname natexlab\endcsname\relax\def\natexlab#1{#1}\fi

\bibitem[{Ahmad et~al.(2021)Ahmad, Peng, and Chang}]{AhmadPC21}
Wasi~Uddin Ahmad, Nanyun Peng, and Kai{-}Wei Chang. 2021.
\newblock {GATE:} graph attention transformer encoder for cross-lingual
  relation and event extraction.
\newblock In \emph{Proceedings of the AAAI Conference on Artificial
  Intelligence}, pages 12462--12470.

\bibitem[{Banerjee et~al.(2018)Banerjee, Moghe, Arora, and
  Khapra}]{banerjee-etal-2018-dataset}
Suman Banerjee, Nikita Moghe, Siddhartha Arora, and Mitesh~M. Khapra. 2018.
\newblock A dataset for building code-mixed goal oriented conversation systems.
\newblock In \emph{Proceedings of the 27th International Conference on
  Computational Linguistics}, pages 3766--3780.

\bibitem[{Bj{\"o}rkelund et~al.(2009)Bj{\"o}rkelund, Hafdell, and
  Nugues}]{bjorkelund-etal-2009-multilingual}
Anders Bj{\"o}rkelund, Love Hafdell, and Pierre Nugues. 2009.
\newblock Multilingual semantic role labeling.
\newblock In \emph{Proceedings of the CoNLL}, pages 43--48.

\bibitem[{Can et~al.(2019)Can, Le, Ha, and Collier}]{can-etal-2019-richer}
Duy-Cat Can, Hoang-Quynh Le, Quang-Thuy Ha, and Nigel Collier. 2019.
\newblock A richer-but-smarter shortest dependency path with attentive
  augmentation for relation extraction.
\newblock In \emph{Proceedings of the 2019 Conference of the North {A}merican
  Chapter of the Association for Computational Linguistics: Human Language
  Technologies}, pages 2902--2912.

\bibitem[{Cucerzan and Yarowsky(1999)}]{cucerzan-yarowsky-1999-language}
Silviu Cucerzan and David Yarowsky. 1999.
\newblock Language independent named entity recognition combining morphological
  and contextual evidence.
\newblock In \emph{Proceedings of the Joint {SIGDAT} Conference on Empirical
  Methods in Natural Language Processing and Very Large Corpora}.

\bibitem[{Culotta and Sorensen(2004)}]{culotta-sorensen-2004-dependency}
Aron Culotta and Jeffrey Sorensen. 2004.
\newblock Dependency tree kernels for relation extraction.
\newblock In \emph{Proceedings of the 42nd Annual Meeting of the Association
  for Computational Linguistics}, pages 423--429.

\bibitem[{Daza and Frank(2019)}]{daza-frank-2019-translate}
Angel Daza and Anette Frank. 2019.
\newblock Translate and label! an encoder-decoder approach for cross-lingual
  semantic role labeling.
\newblock In \emph{Proceedings of the 2019 Conference on Empirical Methods in
  Natural Language Processing and the 9th International Joint Conference on
  Natural Language Processing}, pages 603--615.

\bibitem[{de~Marneffe et~al.(2021)de~Marneffe, Manning, Nivre, and
  Zeman}]{MarneffeMNZ21}
Marie{-}Catherine de~Marneffe, Christopher~D. Manning, Joakim Nivre, and Daniel
  Zeman. 2021.
\newblock Universal dependencies.
\newblock \emph{Comput. Linguistics}, 47(2):255--308.

\bibitem[{Dozat and Manning(2017)}]{DozatM17}
Timothy Dozat and Christopher~D. Manning. 2017.
\newblock Deep biaffine attention for neural dependency parsing.
\newblock In \emph{Proceedings of the 5th International Conference on Learning
  Representations}.

\bibitem[{Fei et~al.(2021)Fei, Li, Li, and Ji}]{FeiGraphSynAAAI21}
Hao Fei, Fei Li, Bobo Li, and Donghong Ji. 2021.
\newblock Encoder-decoder based unified semantic role labeling with label-aware
  syntax.
\newblock In \emph{Proceedings of the AAAI Conference on Artificial
  Intelligence}, pages 12794--12802.

\bibitem[{Fei et~al.(2022)Fei, Wu, Li, Li, Li, Qin, Zhang, Zhang, and
  Chua}]{FeiLasuieNIPS22}
Hao Fei, Shengqiong Wu, Jingye Li, Bobo Li, Fei Li, Libo Qin, Meishan Zhang,
  Min Zhang, and Tat-Seng Chua. 2022.
\newblock Lasuie: Unifying information extraction with latent adaptive
  structure-aware generative language model.
\newblock In \emph{Proceedings of the Advances in Neural Information Processing
  Systems, NeurIPS 2022}, pages 15460--15475.

\bibitem[{Fei et~al.(2020{\natexlab{a}})Fei, Zhang, and
  Ji}]{fei-etal-2020-cross}
Hao Fei, Meishan Zhang, and Donghong Ji. 2020{\natexlab{a}}.
\newblock Cross-lingual semantic role labeling with high-quality translated
  training corpus.
\newblock In \emph{Proceedings of the 58th Annual Meeting of the Association
  for Computational Linguistics}, pages 7014--7026.

\bibitem[{Fei et~al.(2020{\natexlab{b}})Fei, Zhang, Li, and Ji}]{FeiZLJ20}
Hao Fei, Meishan Zhang, Fei Li, and Donghong Ji. 2020{\natexlab{b}}.
\newblock Cross-lingual semantic role labeling with model transfer.
\newblock \emph{{IEEE} {ACM} Trans. Audio Speech Lang. Process.},
  28:2427--2437.

\bibitem[{Gildea and Jurafsky(2000)}]{gildea-jurafsky-2000-automatic}
Daniel Gildea and Daniel Jurafsky. 2000.
\newblock Automatic labeling of semantic roles.
\newblock In \emph{Proceedings of the Annual Meeting of the Association for
  Computational Linguistics}, pages 512--520.

\bibitem[{Halpin and Moore(2006)}]{halpin-moore-2006-event}
Harry Halpin and Johanna~D. Moore. 2006.
\newblock Event extraction in a plot advice agent.
\newblock In \emph{Proceedings of the 21st International Conference on
  Computational Linguistics and 44th Annual Meeting of the Association for
  Computational Linguistics}, pages 857--864.

\bibitem[{hristopher Walker et~al.(2006)hristopher Walker, Strassel, Medero,
  and Maeda}]{ACE5}
hristopher Walker, Stephanie Strassel, Julie Medero, and Kazuaki Maeda. 2006.
\newblock Ace 2005 multilingual training corpus.
\newblock In \emph{Proceedings of Philadelphia: Linguistic Data Consortium}.

\bibitem[{Joshi et~al.(2016)Joshi, Prabhu, Shrivastava, and
  Varma}]{joshi-etal-2016-towards}
Aditya Joshi, Ameya Prabhu, Manish Shrivastava, and Vasudeva Varma. 2016.
\newblock Towards sub-word level compositions for sentiment analysis of
  {H}indi-{E}nglish code mixed text.
\newblock In \emph{Proceedings of the 26th International Conference on
  Computational Linguistics: Technical Papers}, pages 2482--2491.

\bibitem[{Kim et~al.(2010)Kim, Jeong, Lee, and Lee}]{kim-etal-2010-cross}
Seokhwan Kim, Minwoo Jeong, Jonghoon Lee, and Gary~Geunbae Lee. 2010.
\newblock A cross-lingual annotation projection approach for relation
  detection.
\newblock In \emph{Proceedings of the 23rd International Conference on
  Computational Linguistics}, pages 564--571.

\bibitem[{Kozhevnikov and Titov(2013)}]{kozhevnikov-titov-2013-cross}
Mikhail Kozhevnikov and Ivan Titov. 2013.
\newblock Cross-lingual transfer of semantic role labeling models.
\newblock In \emph{Proceedings of the 51st Annual Meeting of the Association
  for Computational Linguistics}, pages 1190--1200.

\bibitem[{Labutov and Lipson(2014)}]{labutov-lipson-2014-generating}
Igor Labutov and Hod Lipson. 2014.
\newblock Generating code-switched text for lexical learning.
\newblock In \emph{Proceedings of the 52nd Annual Meeting of the Association
  for Computational Linguistics}, pages 562--571.

\bibitem[{Lou et~al.(2022)Lou, Gao, Yu, Wang, Zhao, Tu, and Xu}]{LouGYWZTX22}
Chenwei Lou, Jun Gao, Changlong Yu, Wei Wang, Huan Zhao, Weiwei Tu, and Ruifeng
  Xu. 2022.
\newblock Translation-based implicit annotation projection for zero-shot
  cross-lingual event argument extraction.
\newblock In \emph{Proceedings of the 45th International {ACM} {SIGIR}
  Conference on Research and Development in Information Retrieval}, pages
  2076--2081.

\bibitem[{Lu et~al.(2020)Lu, Subburathinam, Ji, May, Chang, Sil, and
  Voss}]{lu-etal-2020-cross}
Di~Lu, Ananya Subburathinam, Heng Ji, Jonathan May, Shih-Fu Chang, Avi Sil, and
  Clare Voss. 2020.
\newblock Cross-lingual structure transfer for zero-resource event extraction.
\newblock In \emph{Proceedings of the Twelfth Language Resources and Evaluation
  Conference}, pages 1976--1981.

\bibitem[{McDonald et~al.(2013)McDonald, Nivre, Quirmbach-Brundage, Goldberg,
  Das, Ganchev, Hall, Petrov, Zhang, T{\"a}ckstr{\"o}m, Bedini,
  Bertomeu~Castell{\'o}, and Lee}]{mcdonald-etal-2013-universal}
Ryan McDonald, Joakim Nivre, Yvonne Quirmbach-Brundage, Yoav Goldberg, Dipanjan
  Das, Kuzman Ganchev, Keith Hall, Slav Petrov, Hao Zhang, Oscar
  T{\"a}ckstr{\"o}m, Claudia Bedini, N{\'u}ria Bertomeu~Castell{\'o}, and
  Jungmee Lee. 2013.
\newblock Universal dependency annotation for multilingual parsing.
\newblock In \emph{Proceedings of the Annual Meeting of the Association for
  Computational Linguistics}, pages 92--97.

\bibitem[{Miwa and Bansal(2016)}]{miwa-bansal-2016-end}
Makoto Miwa and Mohit Bansal. 2016.
\newblock End-to-end relation extraction using {LSTM}s on sequences and tree
  structures.
\newblock In \emph{Proceedings of the 54th Annual Meeting of the Association
  for Computational Linguistics}, pages 1105--1116.

\bibitem[{Mulcaire et~al.(2018)Mulcaire, Swayamdipta, and
  Smith}]{mulcaire-etal-2018-polyglot}
Phoebe Mulcaire, Swabha Swayamdipta, and Noah~A. Smith. 2018.
\newblock Polyglot semantic role labeling.
\newblock In \emph{Proceedings of the Annual Meeting of the Association for
  Computational Linguistics}, pages 667--672.

\bibitem[{Ni and Florian(2019)}]{ni-florian-2019-neural}
Jian Ni and Radu Florian. 2019.
\newblock Neural cross-lingual relation extraction based on bilingual word
  embedding mapping.
\newblock In \emph{Proceedings of the 2019 Conference on Empirical Methods in
  Natural Language Processing and the 9th International Joint Conference on
  Natural Language Processing}, pages 399--409.

\bibitem[{Pad{\'{o}} and Lapata(2009)}]{PadoL09}
Sebastian Pad{\'{o}} and Mirella Lapata. 2009.
\newblock Cross-lingual annotation projection for semantic roles.
\newblock \emph{J. Artif. Intell. Res.}, 36:307--340.

\bibitem[{Qin et~al.(2020)Qin, Ni, Zhang, and Che}]{QinN0C20}
Libo Qin, Minheng Ni, Yue Zhang, and Wanxiang Che. 2020.
\newblock Cosda-ml: Multi-lingual code-switching data augmentation for
  zero-shot cross-lingual {NLP}.
\newblock In \emph{Proceedings of the Twenty-Ninth International Joint
  Conference on Artificial Intelligence}, pages 3853--3860.

\bibitem[{Samanta et~al.(2019)Samanta, Ganguly, and
  Chakrabarti}]{samanta-etal-2019-improved}
Bidisha Samanta, Niloy Ganguly, and Soumen Chakrabarti. 2019.
\newblock Improved sentiment detection via label transfer from monolingual to
  synthetic code-switched text.
\newblock In \emph{Proceedings of the 57th Annual Meeting of the Association
  for Computational Linguistics}, pages 3528--3537.

\bibitem[{Subburathinam et~al.(2019)Subburathinam, Lu, Ji, May, Chang, Sil, and
  Voss}]{subburathinam-etal-2019-cross}
Ananya Subburathinam, Di~Lu, Heng Ji, Jonathan May, Shih-Fu Chang, Avirup Sil,
  and Clare Voss. 2019.
\newblock Cross-lingual structure transfer for relation and event extraction.
\newblock In \emph{Proceedings of the 2019 Conference on Empirical Methods in
  Natural Language Processing and the 9th International Joint Conference on
  Natural Language Processing}, pages 313--325.

\bibitem[{Taghizadeh and Faili(2021)}]{TaghizadehF21}
Nasrin Taghizadeh and Heshaam Faili. 2021.
\newblock Cross-lingual adaptation using universal dependencies.
\newblock \emph{{ACM} Trans. Asian Low Resour. Lang. Inf. Process.},
  20(4):65:1--65:23.

\bibitem[{Taghizadeh and Faili(2022)}]{TaghizadehF22}
Nasrin Taghizadeh and Heshaam Faili. 2022.
\newblock Cross-lingual transfer learning for relation extraction using
  universal dependencies.
\newblock \emph{Comput. Speech Lang.}, 71:101265.

\bibitem[{Velickovic et~al.(2018)Velickovic, Cucurull, Casanova, Romero,
  Li{\`{o}}, and Bengio}]{VelickovicCCRLB18}
Petar Velickovic, Guillem Cucurull, Arantxa Casanova, Adriana Romero, Pietro
  Li{\`{o}}, and Yoshua Bengio. 2018.
\newblock Graph attention networks.
\newblock In \emph{Proceedings of the International Conference on Learning
  Representations}.

\bibitem[{Wu et~al.(2021)Wu, Fei, Ren, Ji, and Li}]{Wu0RJL21}
Shengqiong Wu, Hao Fei, Yafeng Ren, Donghong Ji, and Jingye Li. 2021.
\newblock Learn from syntax: Improving pair-wise aspect and opinion terms
  extraction with rich syntactic knowledge.
\newblock In \emph{Proceedings of the Thirtieth International Joint Conference
  on Artificial Intelligence}, pages 3957--3963.

\bibitem[{Xiao and Guo(2015)}]{xiao-guo-2015-annotation}
Min Xiao and Yuhong Guo. 2015.
\newblock Annotation projection-based representation learning for cross-lingual
  dependency parsing.
\newblock In \emph{Proceedings of the Nineteenth Conference on Computational
  Natural Language Learning}, pages 73--82.

\bibitem[{Zelenko et~al.(2002)Zelenko, Aone, and
  Richardella}]{zelenko-etal-2002-kernel}
Dmitry Zelenko, Chinatsu Aone, and Anthony Richardella. 2002.
\newblock Kernel methods for relation extraction.
\newblock In \emph{Proceedings of the Conference on Empirical Methods in
  Natural Language Processing}, pages 71--78.

\bibitem[{Zhang et~al.(2019)Zhang, Zhang, and Fu}]{zhang-etal-2019-cross}
Meishan Zhang, Yue Zhang, and Guohong Fu. 2019.
\newblock Cross-lingual dependency parsing using code-mixed {T}ree{B}ank.
\newblock In \emph{Proceedings of the Conference on Empirical Methods in
  Natural Language Processing and the 9th International Joint Conference on
  Natural Language Processing}, pages 997--1006.

\bibitem[{Zhang et~al.(2018)Zhang, Qi, and Manning}]{zhang-etal-2018-graph}
Yuhao Zhang, Peng Qi, and Christopher~D. Manning. 2018.
\newblock Graph convolution over pruned dependency trees improves relation
  extraction.
\newblock In \emph{Proceedings of the 2018 Conference on Empirical Methods in
  Natural Language Processing}, pages 2205--2215.

\bibitem[{Zhang et~al.(2021)Zhang, Strubell, and
  Hovy}]{zhang-etal-2021-benefit}
Zhisong Zhang, Emma Strubell, and Eduard Hovy. 2021.
\newblock On the benefit of syntactic supervision for cross-lingual transfer in
  semantic role labeling.
\newblock In \emph{Proceedings of the 2021 Conference on Empirical Methods in
  Natural Language Processing}, pages 6229--6246.

\bibitem[{Zhen et~al.(2021)Zhen, Wang, Fu, Lv, and
  Zhang}]{zhen-etal-2021-chinese}
Ranran Zhen, Rui Wang, Guohong Fu, Chengguo Lv, and Meishan Zhang. 2021.
\newblock {C}hinese opinion role labeling with corpus translation: A pivot
  study.
\newblock In \emph{Proceedings of the 2021 Conference on Empirical Methods in
  Natural Language Processing}, pages 10139--10149.

\end{thebibliography}

\newpage

\appendix

\section{The universal dependency labels}
\label{The universal dependency labels}

In Table \ref{dependency-labels}, we list the dependency labels which are the commonly occurred types.
Please refer to Stanford dependency\footnote{\url{https://nlp.stanford.edu/software/dependencies_manual.pdf}} for more details about the dependency labels.

\begin{table}[!h]
\begin{center}
% \resizebox{1.0\columnwidth}{!}{
\begin{tabular}{ll}
\hline
Dependency Label & Description  \\
\hline
\emph{amod}  & adjectival modifier  \\
\emph{advcl} & adverbial clause modifier\\
\emph{advmod} & adverb modifier  \\
\emph{acomp} & adjectival complement \\
\emph{auxpass} & passive auxiliary \\
\emph{compound} & compound\\
\emph{ccomp} & clausal complement\\
\emph{cc} &  coordination  \\
\emph{conj} &  conjunct   \\
\emph{cop} & copula \\
\emph{det} & determiner \\
\emph{dep} & dependent  \\
\emph{dobj} & direct object	  \\
\emph{mark} & marker \\
\emph{nsubj} & nominal subject\\
\emph{nmod} & nominal modifier \\
\emph{neg} & negation modifier \\
\emph{xcomp} & open clausal complement\\
\hline
\end{tabular}
% }
\end{center}
\caption{The universal dependency labels.
}
\vspace{-3mm}
\label{dependency-labels}
\end{table}

\end{document}